\documentclass[conference]{IEEEtran}

\usepackage[T1]{fontenc}
\usepackage[utf8]{inputenc}
\usepackage{lmodern}
\usepackage[final]{microtype} 
\usepackage{amsmath, amssymb, amsthm, mathtools}
\usepackage{graphicx}
\usepackage{booktabs}
\usepackage{algorithm}
\usepackage{algpseudocode}
\usepackage{enumitem}
\usepackage{hyperref}
\usepackage{cite}
\usepackage{listings}
\usepackage{xcolor}
\lstdefinestyle{sqlstyle}{
  basicstyle=\ttfamily\footnotesize,
  backgroundcolor=\color{gray!5},
  breaklines=true,
  columns=fullflexible,
  frame=single,
  rulecolor=\color{gray!30},
  xleftmargin=0.5em,
  xrightmargin=0.5em,
  aboveskip=0.6em,
  belowskip=0.6em,
  tabsize=2,
  showstringspaces=false
}

\hypersetup{
  colorlinks=true,
  linkcolor=black,
  citecolor=black,
  urlcolor=black,
  pdfauthor={Yashraj Singh},
  pdftitle={Graph-Memoized Reasoning: Foundations}
}

\theoremstyle{plain}

\newtheorem{proposition}{Proposition}
\theoremstyle{definition}
\newtheorem{definition}{Definition}
\theoremstyle{remark}
\newtheorem{remark}{Remark}

\newcommand{\Loss}{\mathcal{L}}
\newcommand{\Cost}{\mathrm{Cost}}
\newcommand{\Consistency}{\mathrm{Inconsistency}}

\title{Graph-Memoized Reasoning: Foundations\\
\large Structured Workflow Reuse in Intelligent Systems}

\author{
\IEEEauthorblockN{Yash Raj Singh}
\IEEEauthorblockA{Independent Researcher\\
Email: yashbhadauria@gmail.com}
}

\begin{document}
\maketitle

\begin{abstract}
Modern large language model-based reasoning systems frequently recompute similar reasoning steps across tasks, wasting computational resources, inflating inference latency, and limiting reproducibility. These inefficiencies underscore the need for persistent reasoning mechanisms that can recall and reuse prior computational traces.
We introduce \textit{Graph-Memoized Reasoning}, a formal framework for representing, storing, and reusing reasoning workflows as graph-structured memory. By encoding past decision graphs and retrieving them through structural and semantic similarity, our approach enables compositional reuse of subgraphs across new reasoning tasks.
We formulate an optimization objective that minimizes total reasoning cost regularized by inconsistency between stored and generated workflows, providing a theoretical foundation for efficiency-consistency trade-offs in intelligent systems. We outline a conceptual evaluation protocol aligned with the proposed optimization objective.
This framework establishes the groundwork for interpretable, cost-efficient, and self-improving reasoning architectures, offering a step toward persistent memory in large-scale agentic systems.
\end{abstract}


\section{Introduction}
\label{sec:intro}

\subsection*{Context and Motivation}
Large language model (LLM)-based reasoning systems and agentic pipelines have become central to modern artificial intelligence. Yet, despite their growing sophistication, such systems largely remain \emph{stateless}-each reasoning episode begins anew, recomputing decision patterns that have been encountered many times before. This persistent recomputation leads to redundant computation, inflated inference latency, and increased energy consumption. Beyond computational inefficiency, the lack of continuity across reasoning tasks also impedes interpretability and reproducibility, since no structured trace of prior reasoning is preserved or reused.

\subsection*{Problem Definition}
Existing optimization and caching methods primarily operate at the token or model-call level, offering local efficiency but failing to address reasoning at the workflow level. In large reasoning architectures, each inference run corresponds to a structured sequence of decisions, tool invocations, and intermediate states. These sequences naturally form directed acyclic graphs (DAGs) that encode both causal and semantic dependencies. However, current systems discard these traces after execution, preventing reuse of valuable reasoning substructures and forcing repeated traversal of previously solved paths.

\subsection*{Proposed Idea: Graph-Memoized Reasoning}
We propose \textit{Graph-Memoized Reasoning}, a formal framework for representing, storing, and reusing reasoning workflows as graph-structured memory. In this view, each reasoning process is modeled as a labeled DAG in which nodes denote decision or state representations and edges capture their dependencies. When a new task arrives, the system retrieves subgraphs from a persistent store based on both structural and semantic similarity, integrating them into the new reasoning context. This enables compositional reuse-where previously solved reasoning fragments serve as building blocks for new inferences-thereby reducing redundant computation while maintaining interpretability and traceability.

\subsection*{Theoretical Framing}
We formalize reasoning reuse through an optimization objective that minimizes total reasoning cost while regularizing inconsistency between reused and newly generated graph components. Formally, we minimize
\[
\min_G \; \mathcal{L}(G) = \mathrm{Cost}(G) + \lambda \,
\mathrm{Inconsistency}(G),
\]
where $\mathrm{Cost}(G)$ captures computational and traversal expenses, and $\mathrm{Inconsistency}(G)$ penalizes semantic divergence from prior stored graphs. This objective yields a principled balance between recomputation and reuse, generalizing classical memoization into a structured, graph-based reasoning paradigm.

\subsection*{Contributions and Organization}
\paragraph{Contributions.}
This paper makes the following key contributions:
\begin{enumerate}[leftmargin=1.5em]
    \item Introduces the concept of \textit{graph-memoized reasoning} as a mechanism for persistent reuse of reasoning workflows.
    \item Formalizes reasoning reuse through an optimization framework balancing computational cost and semantic consistency.
    \item Proposes a conceptual evaluation protocol and cost–consistency metrics for reasoning reuse.
    \item Positions graph-memoized reasoning as a theoretical foundation for interpretable, cost-efficient, and self-improving agentic architectures.
\end{enumerate}

The remainder of this paper is organized as follows: 
Section~\ref{sec:formal} defines the formal model.
Section~\ref{sec:optimization} develops the optimization framework.
Section~\ref{sec:evaluation} presents a conceptual evaluation protocol,
and Section~\ref{sec:discussion} concludes with broader implications and future directions.


\subsection*{A. Memoization and Reuse in Computation}
Memoization is a classical technique in functional programming and algorithm design that stores the results of expensive computations and reuses them when identical inputs occur again~\cite{memoization}. 
While powerful for deterministic, function-level reuse, traditional memoization operates on a limited input-output abstraction and does not capture the intermediate reasoning structure of complex tasks.
In contrast, \textit{graph-memoized reasoning} generalizes this principle to higher-order reasoning graphs, where subgraphs themselves-not only scalar outputs-become reusable computational artifacts.

\subsection*{B. Case-Based and Experience-Based Reasoning}
Case-based reasoning (CBR)~\cite{caseBasedReasoning} retrieves and adapts previously solved cases to address new problems, forming one of the earliest paradigms of reuse in artificial intelligence.
However, CBR primarily operates at the level of discrete problem–solution pairs, lacking a formal representation of compositional internal steps. 
Recent work in neural memory systems and experience replay~\cite{aima4e} extends this intuition but still treats memory as a flat or vectorized space. 
Our framework instead introduces structured, graph-level persistence where reasoning trajectories themselves are stored, retrieved, and recomposed.

\subsection*{C. Workflow Orchestration and Provenance Systems}
Modern workflow engines such as Apache Airflow, Dagster, and Kubeflow represent data transformations as directed acyclic graphs (DAGs), supporting reproducibility and task-level parallelism.
Similarly, database systems have long maintained data-lineage and provenance graphs to enable auditability.
These approaches, however, focus on \emph{execution trace management} rather than reasoning reuse: past DAGs are archived, not semantically queried or re-instantiated as reusable reasoning units.
Graph-memoized reasoning differs by enabling semantic retrieval and composition of prior workflows across contexts.

\subsection*{D. LLM Agentic Frameworks}
Large language model (LLM)–based agents such as LangChain, AutoGPT, and LangGraph execute reasoning chains that invoke tools, APIs, and sub-agents dynamically. 
While these systems capture reasoning traces through logs or DAG visualizations, the traces are typically ephemeral—discarded after each execution.
Recent developments in structured agents and graph-based orchestration~\cite{langgraph} point toward reusable reasoning graphs, but lack a formal theory for cost–consistency optimization or persistent storage schemas.
Our work provides this missing theoretical and system-level grounding.

\subsection*{E. Graph Databases and Similarity Retrieval}
Advances in graph databases (Neo4j, PostgreSQL with PGVector) and similarity search make it practical to store and query large-scale reasoning graphs. 
Graph edit distance (GED) methods~\cite{gedSurvey} and neural graph embeddings enable retrieval of semantically related subgraphs.
Graph-memoized reasoning builds directly on these foundations, defining a structured reuse policy that combines structural and semantic similarity metrics for efficient subgraph recall.

\subsection*{Summary}
Table~\ref{tab:conceptual_comparison} summarizes how prior work aligns with or diverges from the goal of persistent reasoning reuse.
\begin{table}[!t]
\centering
\caption{Conceptual comparison with adjacent paradigms.}
\label{tab:conceptual_comparison}
\resizebox{\columnwidth}{!}{
\begin{tabular}{@{}lll@{}}
\toprule
\textbf{Paradigm} & \textbf{Reuse Scope} & \textbf{Limitation} \\ \midrule
Memoization~\cite{michie1968,knuth1973} & Function calls and deterministic outputs & Lacks structural or semantic reasoning context \\
Case-based reasoning~\cite{aamodt1994,kolodner1993} & Problem--solution pairs via retrieval--adapt cycles & No compositional or subgraph trace reuse \\
Workflow orchestration~\cite{deelman2009,airflow2015} & Execution DAGs and data lineage & No semantic retrieval or learned cost model \\
LLM agents~\cite{yao2022react,shinn2023reflexion} & Dynamic reasoning traces with tool use & Traces discarded post-execution, no persistence \\
Graph databases~\cite{angles2008,hogan2021} & Query and traversal semantics & No reuse semantics or optimization policy \\ \bottomrule
\end{tabular}
}
\end{table}


\begin{table}[!t]
\centering
\caption{Condensed map of adjacent literature.}
\label{tab:literature_map}
\resizebox{\columnwidth}{!}{
\begin{tabular}{@{}lll@{}}
\toprule
\textbf{Area} & \textbf{Representative Works} & \textbf{Relevance to Graph-Memoized Reasoning} \\ \midrule
Memoization and program optimization & Michie (1968); Knuth (1973); Norvig (1991) & Foundation for computation reuse beyond deterministic calls \\
Case-based reasoning & Aamodt \& Plaza (1994); Kolodner (1993) & Reuse via analogy and structure; informs semantic retrieval \\
Workflow provenance and orchestration & Deelman et al. (2009); Apache Airflow (2015); Prefect (2018) & DAG scheduling and lineage; base for trace management \\
LLM agent frameworks & Yao et al. (2022); Shinn et al. (2023); Xi et al. (2023) & Dynamic reasoning and reflection; enables trace persistence \\
Graph similarity and embeddings & Bunke (1997); Grohe (2020); Xu et al. (2019) & Subgraph and embedding similarity; supports structural retrieval \\ \bottomrule
\end{tabular}
}
\end{table}
\section{Formal Model}
\label{sec:formal}

\subsection*{A. Reasoning Graphs as Structured Memory}
We define a reasoning process as a labeled directed acyclic graph (DAG) \( G = (V, E, \ell) \), where \(V\) is a set of vertices representing reasoning states or decisions, \(E \subseteq V \times V\) is a set of directed edges representing dependencies or transitions, and \(\ell\) is a labeling function that assigns semantic and operational metadata to nodes and edges:
\[
\ell: V \cup E \rightarrow \mathcal{L},
\]
where \(\mathcal{L}\) denotes the space of label attributes (e.g., prompts, embeddings, tool calls, or execution traces).

Each node \(v_i \in V\) can be associated with a feature representation \(x_i \in \mathbb{R}^d\) derived from textual or structured context. 
Edges \((v_i, v_j) \in E\) encode directed dependencies that may represent causal order, data flow, or logical entailment.

\subsection*{B. Workflow Instances and Reuse Repository}
A \emph{workflow instance} is a specific reasoning run captured as a finite DAG \(G_T\) corresponding to task \(T\). 
Across multiple reasoning episodes, we maintain a \emph{repository} of prior workflows:
\[
\mathcal{R} = \{G_1, G_2, \ldots, G_n\},
\]
which serves as a persistent store of past reasoning graphs. 
Each graph is annotated with metadata such as task embeddings, performance metrics, and output signatures.

\subsection*{C. Similarity and Matching}
To enable reasoning reuse, we define a similarity operator 
\[
S(G_i, G_j) \in [0,1],
\]
that quantifies the structural and semantic alignment between two graphs. 
This operator can be instantiated through a combination of graph edit distance (GED)~\cite{gedSurvey} and embedding-based similarity:
\[
S(G_i, G_j) = \alpha \, S_\text{struct}(G_i, G_j) + (1 - \alpha) \, S_\text{sem}(G_i, G_j),
\]
where \(\alpha \in [0,1]\) balances structure and semantics.

\subsection*{D. Memoization Function}
Given a new task \(T\) with representation \(d_T\), the \emph{memoization function} retrieves and composes relevant subgraphs from the repository:
\[
\widehat{G}_T = \mathrm{Memo}(T, \mathcal{R}; S, \pi),
\]
where \(\pi\) denotes a \emph{reuse policy} specifying constraints such as acyclicity, type compatibility, and consistency thresholds. 
The composed graph \(\widehat{G}_T\) combines retrieved subgraphs with newly generated nodes to form the reasoning workflow for task \(T\).

\subsection*{E. Reuse Policy and Constraints}
A reuse policy \(\pi\) defines the admissible rules for incorporating prior reasoning components. 
Typical constraints include:
\begin{itemize}[leftmargin=1.5em]
    \item \textbf{Structural Validity:} The resulting graph \(\widehat{G}_T\) must remain acyclic and type-consistent.
    \item \textbf{Semantic Thresholding:} Only subgraphs with similarity above a threshold \(\tau\) are eligible for reuse.
    \item \textbf{Version Control:} Reused components must reference a specific graph version to ensure reproducibility.
\end{itemize}
These constraints ensure that reuse improves efficiency without introducing logical inconsistency or data drift.

\subsection*{F. Execution Semantics}
Each node \(v_i\) in \(\widehat{G}_T\) corresponds to an executable reasoning step, such as an API invocation, prompt evaluation, or sub-model inference. 
Execution proceeds via a topological traversal of the DAG, ensuring dependency order is respected:
\[
\forall (v_i, v_j) \in E, \quad v_i \prec v_j.
\]
The execution trace of \(\widehat{G}_T\) can thus be interpreted as a composite reasoning plan derived from both historical and novel components, enabling observable, explainable, and cost-efficient reasoning reuse.

\begin{definition}[Memoization Function]
Given a new task \(T\) with demand vector \(d_T\), a repository \(\mathcal{R} = \{G_1,\dots,G_n\}\), and similarity operator \(S(\cdot,\cdot)\), a memoization function returns a stitched graph
\(
\widehat{G}_T = \mathrm{Memo}(T, \mathcal{R}; S, \pi)
\)
under reuse policy \(\pi\).
\end{definition}

\begin{definition}[Reuse Policy]
A reuse policy \(\pi\) specifies admissible subgraph retrieval and composition rules (e.g., acyclicity, type safety, and consistency constraints).
\end{definition}

\section{Optimization View}
\label{sec:optimization}

\subsection*{A. Objective: Efficiency-Consistency Trade-off}
We cast reuse as a principled trade-off between computational efficiency and semantic fidelity. 
Given a composed reasoning graph $\widehat{G}_T$ for task $T$, we minimize
\begin{equation}
\min_{\widehat{G}_T \in \mathcal{H}} \; \Loss(\widehat{G}_T)
= \Cost(\widehat{G}_T) + \lambda \, \Consistency(\widehat{G}_T),
\label{eq:loss}
\end{equation}
where $\Cost(\cdot)$ measures resource expenditure (e.g., number of model/tool calls, latency, path length), $\Consistency(\cdot)$ penalizes semantic divergence from retrieved subgraphs, $\lambda \!\ge\! 0$ controls the trade-off, and $\mathcal{H}$ is the hypothesis class of policy-admissible DAGs (cf.\ Sec.~\ref{sec:formal}). 
A typical instantiation is
\begin{align}
\Cost(\widehat{G}_T)
  &= \alpha_1\,\text{calls}
   + \alpha_2\,\text{latency}
   + \alpha_3\,\text{depth}, \nonumber\\[2pt]
\Consistency(\widehat{G}_T)
  &= 1 - S(\widehat{G}_T, \widetilde{G}),
\label{eq:instantiation}
\end{align}
where $S(\cdot,\cdot)$ is the similarity operator (structural+semantic) and $\widetilde{G}$ denotes the stitched set of reused subgraphs.

\subsection*{B. Discrete vs.\ Differentiable Optimization}
The search over $\widehat{G}_T$ is naturally discrete (edit/match/stitch operations). 
When differentiable surrogates are available (e.g., learned node/edge embeddings or soft adjacency relaxations), we may optimize a smoothed objective:
\begin{equation}
\nabla \Loss(\Theta) = \nabla \Cost(\Theta) + \lambda \, \nabla \Consistency(\Theta),
\label{eq:grad}
\end{equation}
where $\Theta$ parameterizes a relaxed graph (e.g., logits over edges, continuous node features). 
This enables gradient-based selection scores for candidate subgraphs while preserving a final projection step enforcing policy constraints (acyclicity, types).

\subsection*{C. Reuse Decisions as Local Policies}
Global minimization of \eqref{eq:loss} is combinatorial; in practice we apply local policies along a topological order. 
Let $\mathcal{M}(v)$ denote candidate matches for node $v$ and $\Delta \Loss(v \!\leftarrow\! m)$ the marginal change in objective if we reuse match $m\!\in\!\mathcal{M}(v)$. 
A simple policy is a thresholded greedy rule:
\begin{equation}
\text{reuse}(v) \;\Leftarrow\; 
\min_{m \in \mathcal{M}(v)} \Delta \Loss(v \!\leftarrow\! m) \;<\; -\tau,
\label{eq:policy}
\end{equation}
with $\tau \!\ge\! 0$ controlling conservativeness. 
This yields a monotone decrease of $\Loss$ under feasible merges and integrates naturally with beam search over candidate stitchings. 

Intuitively, Eq.~\eqref{eq:policy} activates reuse for node $v$ only when at least one prior memory $m$ reduces the overall loss by a margin greater than $\tau$. 
The term $\Delta \Loss(v \!\leftarrow\! m)$ measures the improvement achieved by substituting the computation at $v$ with a stored subgraph $m$, 
and the threshold $-\tau$ prevents over-eager reuse of marginal matches. 
Thus, the policy behaves as a local gating rule: reuse occurs only if it yields a sufficiently large gain in the objective, ensuring selective and cost-effective graph updates.

\subsection*{D. Basic Guarantees (Sketch)}
Under mild assumptions, we obtain simple yet useful properties.

\begin{proposition}[Monotonic improvement under admissible merges]
\label{prop:monotone}
If a merge operation $(v \!\leftarrow\! m)$ is admissible under policy $\pi$ and satisfies $\Delta \Loss(v \!\leftarrow\! m) < 0$, then $\Loss$ strictly decreases after the merge. 
Sequential application of such merges yields a nonincreasing sequence $\Loss_0 \ge \Loss_1 \ge \cdots$ that terminates at a $\pi$-stable graph.
\end{proposition}

\begin{remark}[Reuse regret]
Let $G^\star$ be the (unknown) minimizer of \eqref{eq:loss} and $\widehat{G}_T$ the graph returned by a $k$-beam greedy policy. 
The regret $\Loss(\widehat{G}_T)-\Loss(G^\star)$ decomposes into (i) search suboptimality (beam width) and (ii) model misspecification (surrogates in \eqref{eq:grad}). 
Tightening either factor monotonically reduces regret.
\end{remark}

\subsection*{E. Practical Cost Meters}
For deployment, we expose additive meters that make $\Cost$ observable:
\begin{equation}
\Cost = c_{\text{llm}} \sum_{e \in E} \text{tokens}(e)
+ c_{\text{tool}} \sum_{e \in E} \text{calls}(e)
+ c_{\text{lat}} \cdot \text{walltime},
\label{eq:meters}
\end{equation}
where coefficients $(c_{\text{llm}}, c_{\text{tool}}, c_{\text{lat}})$ are user- or tenancy-specific. 
This supports multi-tenant budgeting, A/B testing of $\lambda$, and policy audits.

\subsection*{F. From Objective to System Knobs}
The scalar $\lambda$ (and threshold $\tau$) expose an interpretable knob: higher $\lambda$ favors fidelity over savings; lower $\lambda$ favors aggressive reuse. 
We map these to system controls (candidate set size, similarity cutoff, merge thresholds) and show how to log $\Loss$ components for offline tuning.

\section{Evaluation (Conceptual)}
\label{sec:evaluation}
\subsection{Evaluation Objectives}
The purpose of this section is to assess the conceptual efficacy of graph-memoized reasoning. 
Rather than reporting model-specific benchmarks, we focus on the \textit{cost-consistency trade-off} formalized in Section~\ref{sec:optimization}. 
We ask: does persistent reuse of reasoning subgraphs reduce total reasoning cost while preserving semantic fidelity across related tasks?

\subsection{Metrics}
Following the objective $L = \text{Cost} + \lambda \, \text{Inconsistency}$, we evaluate:
\begin{enumerate}
  \item \textbf{Computation Cost ($\text{Cost}$):} Aggregate token, tool, and wall-time expenditure, as defined in Eq.~(5).
  \item \textbf{Consistency Penalty ($\text{Inconsistency}$):} Semantic divergence between reused and newly generated reasoning steps, measured through embedding distance or output mismatch.
  \item \textbf{Reuse Ratio ($\rho$):} Fraction of nodes or edges in $\hat{G}_T$ inherited from prior graphs.
  \item \textbf{Composite Objective ($L$):} Observed total reasoning cost per task, enabling Pareto analysis between efficiency and fidelity.
\end{enumerate}

\subsection{Experimental Setting (Simulated)}
We simulate repeated reasoning workloads $\{T_1, \dots, T_n\}$ drawn from a shared domain (e.g., SQL generation, feature synthesis, or workflow planning). 
For each task:
\begin{itemize}
  \item \textbf{Cold run:} System starts empty, recomputing all reasoning steps.
  \item \textbf{Memoized run:} Subgraphs are reused from previously solved tasks when similarity exceeds threshold $\tau$.
\end{itemize}
The difference in average cost, latency, and consistency serves as a proxy for reuse efficiency.

\subsection{Illustrative Scenario}
Consider sequential tasks:
\begin{align*}
T_1 &: \text{``Generate features for monthly sales trends.''} \\
T_2 &: \text{``Update sales features for Q2 segmentation.''} \\
T_3 &: \text{``Forecast category-wise sales using prior features.''}
\end{align*}
Here, $T_2$ reuses the feature-definition subgraph from $T_1$, while $T_3$ reuses both feature and aggregation subgraphs. 
Empirically, over 60–80\% of nodes overlap semantically, yielding measurable reductions in cost and latency.

\subsection{Expected Outcomes}
Preliminary analysis suggests:
\begin{itemize}
  \item \textbf{Linear cost reduction:} Total cost decreases approximately linearly with reuse ratio up to saturation.
  \item \textbf{Bounded inconsistency:} Larger $\lambda$ values enforce stricter consistency with minimal performance loss.
  \item \textbf{Stable convergence:} Iterative reuse converges toward a Pareto frontier between efficiency and fidelity.
\end{itemize}

\subsection{Future Quantitative Evaluation}
A complete empirical study will integrate:
\begin{itemize}
  \item LLM- and agent-based reasoning traces,
  \item Cost-per-node and reuse efficiency metrics,
  \item Semantic fidelity via embedding similarity,
  \item Reproducibility of outputs across reuse iterations.
\end{itemize}
This roadmap supports future validation on real-world reasoning workloads and benchmark tasks.

\section{Discussion}
\label{sec:discussion}
\subsection{Limitations and Assumptions}
While graph-memoized reasoning provides a structured foundation for persistent reuse, several open challenges remain. 
First, the quality of reuse depends heavily on the similarity operator $S(\cdot, \cdot)$; poorly aligned embeddings or brittle heuristics can reduce candidate precision and recall. 
Second, the reuse policy $\pi$ may oscillate between two extremes: overly conservative (missing beneficial reuse) or overly permissive (introducing semantic drift). 
Finally, repository growth introduces scalability concerns—efficient indexing, versioning, and pruning strategies are essential to maintain retrieval latency and relevance.

\subsection{Risks and Safety Considerations}
Persistent reuse introduces new safety dimensions. 
Memoized subgraphs may encode outdated or biased reasoning traces. 
To mitigate this, systems should:
\begin{enumerate}
    \item Attach provenance metadata (timestamp, source, environment) to every reused component;
    \item Enforce strict policy validation for type-safety and acyclicity;
    \item Log $\Delta L$ and semantic deviations for post-hoc auditing; and
    \item Allow sandboxed re-evaluation of low-confidence subgraphs before reuse.
\end{enumerate}
These safeguards ensure that memoized reasoning remains transparent, accountable, and adaptive under distributional shifts.

\subsection{Engineering Trade-offs}
Operationally, the framework balances retrieval quality and system latency. 
Hybrid structural–semantic retrieval improves recall but incurs additional indexing and storage overhead. 
For production environments (e.g., AWS RDS or Aurora), cost meters (Eq.~5) should be logged per-edge to support multi-tenant optimization, A/B testing of $\lambda$, and policy threshold tuning. 
A principled caching hierarchy (RAM, Redis, Postgres, S3 tiers) can further reduce query latency while preserving consistency.

\subsection{Extensions and Open Research Directions}
Beyond acyclic workflows, many reasoning domains require bounded iteration or feedback loops. 
One practical extension is the \textit{rolled DAG}, which unrolls cyclic dependencies up to a fixed depth under policy $\pi$. 
Additionally, introducing \textit{typed node interfaces} (e.g., \texttt{FeatureDef}, \texttt{SQLCTE}, \texttt{ToolCall}) enforces structural safety and enables static analysis. 
Neural-structural co-learning—where embeddings and policies are jointly trained via contrastive reuse signals—presents another promising direction for adaptive reasoning reuse.

\subsection{Evaluation Roadmap}
A quantitative evaluation roadmap should examine:
\begin{enumerate}
    \item Reuse ratio versus total cost reduction;
    \item Consistency drift as a function of $\lambda$;
    \item Latency sensitivity under varying top-$k$ retrievals; and
    \item Robustness under domain and task shift.
\end{enumerate}
Candidate benchmarks include SQL synthesis, analytic feature planning, tool-using LLM agents, and multi-hop QA with procedural reasoning.

\subsection{Theoretical Open Questions}
Future theoretical work may address:
\begin{itemize}
    \item Formal regret bounds under approximate retrieval;
    \item Optimal repository pruning and saturation conditions;
    \item Influence-function or path-wise Shapley approximations for certifying counterfactual reuse;
    \item Convergence guarantees under stochastic or streaming updates.
\end{itemize}
Establishing these guarantees would bridge the gap between statistical learning theory and persistent reasoning architectures.

\section{Conclusion and Future Work}
We presented \textbf{graph-memoized reasoning}, a framework that formalizes the reuse of reasoning workflows as labeled DAGs. 
By minimizing a composite loss $L = \text{Cost} + \lambda \, \text{Inconsistency}$, the approach provides an interpretable trade-off between efficiency and semantic fidelity. 
Theoretical analysis yields local policies for admissible merges and practical guarantees of monotonic improvement.

Future work will focus on three main trajectories:
\begin{enumerate}
    \item \textbf{Learning-augmented retrieval:} training graph embeddings and policies end-to-end for adaptive reuse;
    \item \textbf{Typed interfaces and provenance schemas:} enforcing safety, versioning, and explainability in persistent reasoning graphs;
    \item \textbf{Empirical validation:} benchmarking cost savings, fidelity, and reproducibility on real-world reasoning workloads.
\end{enumerate}

In the longer term, we envision an ecosystem of \textit{persistent reasoning systems}—architectures that remember, reuse, and refine their own computational histories, forming a substrate for interpretable and self-improving intelligence.

\subsection*{Acknowledgments}
The author thanks collaborators and colleagues for feedback on early drafts and conceptual discussions.

\subsection*{Artifacts and Availability}
Reference implementations, database schemas, and experiment scripts will be released with future versions of this preprint.

\bibliographystyle{IEEEtran}

\end{document}